\documentclass[twoside,11pt]{article}

\usepackage{jmlr2e}
\usepackage{amsmath}
\usepackage{xspace}
\usepackage[dvipsnames]{xcolor}
\usepackage[nodisplayskipstretch]{setspace}
\usepackage{numprint}
\usepackage{algorithm2e}
\let\oldnl\nl
\newcommand{\nonl}{\renewcommand{\nl}{\let\nl\oldnl}}
	\newcommand{\assign}{\ensuremath{:=}}
\usepackage[T1]{fontenc}
\usepackage{transparent}
\usepackage[latin1]{inputenc}
\usepackage{xcolor}
\usepackage{amsmath}
\usepackage{xcolor}
\usepackage{amssymb}
\usepackage{amsfonts}
\usepackage[mathscr]{euscript}
\usepackage{graphicx}	
\usepackage{caption}
\usepackage{multirow}
\usepackage{hyperref}
\usepackage{subcaption}
\usepackage{array}
\usepackage{pdflscape}
\makeatletter
\newcolumntype{P}[1]{>{\centering\arraybackslash}p{#1}}
\newcommand{\thickhline}{%
    \noalign {\ifnum 0=`}\fi \hrule height 1pt
    \futurelet \reserved@a \@xhline
}
\newcolumntype{"}{@{\hskip\tabcolsep\vrule width 1pt\hskip\tabcolsep}}
\newcommand{\N}{\ensuremath{\mathcal{N}}}
\newcommand{\Prod}{\ensuremath{\mathcal{P}}}

\newcommand{\occ}{\text{occ}}
\newcommand{\algname}{$\texttt{NRGreedy}_\texttt{fix}$\xspace}

\ShortHeadings{The Generalized Smallest Grammar Problem}{P. Siyari, M. Gall\'e}
\firstpageno{1}

\begin{document}

\title{The Generalized Smallest Grammar Problem}

\author{\name Payam Siyari \email payamsiyari@gatech.edu \\
      \addr College of Computing, 
      Georgia Institute of Technology\\
      Atlanta, GA 30332, USA
       \AND
       \name Matthias Gall\'e \email matthias.galle@xrce.xerox.com \\
       \addr Xerox Research Centre Europe\\
       Meylan, France}

\editor{}

\maketitle

\begin{abstract}
The Smallest Grammar Problem -- the problem of finding the smallest context-free grammar that generates exactly one given sequence -- has never been successfully applied to grammatical inference.
We investigate the reasons and propose an extended formulation that seeks to minimize non-recursive grammars, instead of straight-line programs. 
In addition, we provide very efficient algorithms that approximate the minimization problem of this class of grammars.
Our empirical evaluation shows that we are able to find smaller models than the current best approximations to the Smallest Grammar Problem on standard benchmarks, and that the inferred rules capture much better the syntactic structure of natural language.
\end{abstract}


\section{Introduction}

The Smallest Grammar Problem (SGP) is the optimization problem of finding a smallest context-free grammar that generates exactly a given sequence.
As such, it has some superficial resemblance to Grammatical Inference, both because of the choice of model to structure the data (a formal grammar) and because the goal of identifying structure is explicitly called-out as a potential application of SGP and its extensions~\citep{Nevill1997,Charikar2005,Galle2011,Siyari2016}.
However, concrete applications so far of the SGP to grammatical inference are either remarkably absent in the literature, or have been reported to utterly fail~\citep{Eyraud2006}.
The main reason for this is that the definition of the SGP limits the inferred models to be straight-line programs which have no generalization capacity. We present here an alternative definition which removes this constraint and design an efficient algorithm that achieves at the same time:
\begin{itemize}
	\item smaller grammars than the state-of-the-art, measured on standard benchmarks.
    \item generalized rules which correspond to the true underlying syntactic structure, measured on the Penn Tree bank.
\end{itemize}

We achieve this by extending the formalism from simple straight-line grammars to non-recursive grammars.
Our algorithm takes as input any straight-line grammar and infers additional generalization rules, optimizing a score function inspired both by the distributional hypothesis~\citep{Harris1954} and regularizing through the Minimum Description Length (MDL) principle.
It is very efficient in practice and can easily be run on sequences of millions of symbols.

\section{Related Work}
\label{sect:related}
We take inspiration from three related areas:  the work around the Smallest Grammar Problem, concrete implementations of Harris substitutability theory~\citep{Harris1954} and the use of Minimum Description Length principle in grammatical inference.

The \textbf{Smallest Grammar Problem} was formally introduced and analyzed in~\citet{Charikar2005}, and provides a theoretical framework for the popular Sequitur algorithm~\citep{Nevill1997}, as well as the work around grammar-based compression~\citep{Kieffer2000}.
It is defined as the combinatorial problem of finding a smallest context-free grammar that generates exactly the given input sequence.
By reducing the expression power from Turing machines to context-free grammars it becomes a computable (although intractable) version of Kolmogorov Complexity.
This relationship is also reflected in the interest of studying that problem: by finding smaller grammars, the hope is not only to find better compression techniques but also to model better the redundancies and therefore the structure of the sequence.
The current state-of-the-art algorithms that obtain the smallest grammars on standard benchmarks are based on a search space introduced in~\citet{Carrascosa2011} which decouples the search of the optimal set of substrings to be used, and the choice of how to combine these in an optimal parsing of the input sequence.
That parsing can be solved optimally in an efficient way, and the algorithms diverge in how they navigate the space of possible substrings to be used, trading off efficiency with a broader search.
These algorithms include IRRMGP~\citep{Carrascosa2011b} -- a greedy algorithm --, ZZ~\citep{Carrascosa2010} -- a hill-climbing approach --, and MMAS-GA~\citep{Benz2013} -- a genetic algorithm --.
While their worst-case approximation ratio is hard to analyze, they perform empirically better than other algorithms with theoretical guarantees.
Applications to grammar learning has been studied in~\citet[Chapter~3]{Eyraud2006}, which concludes that the failure to retrieve meaningful structure is due to the fact that \texttt{Sequitur} ``est un algorithme dont le but est de compresser un texte \`a' l'aide d'une grammaire  [$\dots$] Or la  fr\'equence d'apparition d'un motif n'est pas une mesure permettant de savoir si ce motif est un constituant''\footnote{``is an algorithm whose goal is to compress a text through a grammar [$\dots$] but the frequency of occurrences of a motif is not a signal to decide whether a motif is a constituent''}.

\smallskip

Algorithms to find smaller grammars focus on \textit{intrinsic} properties of substrings (occurrences, length).
However, in grammatical inference a primary focus is the \textit{context} in which a substring occurs in, a principle that traces back to Harris \textbf{substitutability theory}~\citep{Harris1954}.
Different classes of substituable languages have been defined, which all start from the intuition that if two words occur in the same context they should belong to the same semantic class (be generated by the same non-terminal).
Occurrences of strings $uwv$ and $uw'v$ are therefore a signal that the words $w$ and $w'$ are substituable one by the other (as in ``The car is fast'' and ``The bike is fast'', with $w=\text{car}$ and $w'=\text{bike}$).
In recent years, very good learnability results have been obtained with different variations of substituable languages~\citep{Yoshinaka2008,Clark2007,Luque2010,Coste2012}, and those insights have long been the basis of unsupervised learning algorithms applied to natural language text \citep{ABL,ADIOS,Scicluna2014}.

\smallskip

The \textbf{Minimum Description Length} principle is a popular approach for model selection, and states that the best model to describe some data is the one that minimizes jointly the size of the model and the cost of describing the data given the model.
This has been applied often in tools targeted to discover meaningful substructure, including grammatical inference~\citep{Cook1994,Keller1997}.
In this context, the grammars obtained through the SGP are learning the data by heart as they do not perform any generalization.
Instead of this, we propose to extend this model and to control its generalization capacity through MDL.
This point is also our main divergence with \citet{Chirathamjaree1980} who also infer non-recursive grammars (although only linear grammars). 
While their use of the edit distance to decide which rules to re-use reflects the minimization intuition in MDL, this is not used when deciding when to generalize. 
It is therefore not straightforward how to measure the cost of deriving a particular string.
Without such a cost, any comparison with algorithms inferring straight-line grammars seems unfair.
In this paper, much of our attention is focused on how to derive non-recursive grammars that still are able to generate the input sequence exactly.
While such a purist strategy may not be ideal to obtain the best empirical performance for structure inference, in this line of work we want to test the boundaries of such an approach and to be able to make fair comparisons with the results coming from the SGP.

\section{Model}
By focusing on grammars that generate a single sequence only, the SGP limits itself to straight-line grammars, which are context-free grammars which do neither branch nor loop.
We propose here to relax the first of these constraints, allowing branching non-terminals.
Such ``non-recursive grammars'' have found use in natural-language processing, where several applications have this characteristic~\citep{Nederhof2002}, despite the fact that they are only able to generate finite languages. 

\begin{definition}
A non-recursive context free grammar is a tuple $G = \langle \Sigma, \N, S, \Prod \rangle$, with terminals $\Sigma$, non-terminals $\N$ starting symbol $S$ and context-free production rules $\Prod$ such that for any $A,B \in \N$ not necessarily distinct, if $B$ occurs in a derivation of $A$, then $A$ does not occur in a derivation of $B$.
\end{definition}

Different from straight-line grammars, the language generated by non-recursive CFG can be larger than a single string, although it is always finite.
In the spirit of the smallest grammar problem, we are still interested in encoding exactly one given string and have therefore to specify which of all the strings in the language is the encoded one.

The size of such a grammar with respect to a specific sequence $s$ -- which we will then try to minimize -- will therefore be the sum of two factors: the size of the general grammar, and the cost of specifying $s$:

\begin{definition}
	Given a non-recursive context-free grammar $G = \langle \Sigma, \N, S, \Prod \rangle$, the size of $G$ wrt $s$ is defined as:
    
$$ \displaystyle   |G|_s = \sum_{N \rightarrow \alpha \in \Prod} \left( |\alpha| + 1 \right)  +  \text{cost}(s | G)$$ 
\end{definition}

\noindent where $\textit{cost(s|G)}$ is the cost of expressing $s$ given $G$, and should be expressed in the same unit that the grammar itself, namely symbols.

We are now ready to define our generalized version of the smallest grammar problem:

\begin{definition}
Given a sequence $s$, a smallest non-recursive grammar is a non-recursive context-free grammar $G$ such that $s \in L(G)$ and $|G|_s$ is minimal.
\end{definition}

Note that we do not put any restrictions on the language that the grammar could generate.
A more general grammar (one that generates a larger language) may have less or shorter production rules, but the decoding may be more expensive.
This is a standard trade-off in any MDL formulation.
In the extreme case, where $|L(G)|=1$ and $\text{cost}(s|G)=0$, this reduces to the traditional smallest grammar problem.
The goal of generalizing the definition is that by adding ambiguities, the grammar would end up having smaller size even with some amount of cost being added for resolving the ambiguities.

Before we define $\textit{cost}(s | G)$, we need to introduce the mentioned type of ambiguity that when introduced, is potential to lead us to smaller grammars. 
From a compression perspective, we are interested in capturing more flexible patterns than just exact repeats.
In that way, we are looking for non-terminals that generate words $v$ and $v'$, where both words are different although \textit{similar}, so that disambiguating between them is cheap.
Several such similarities have been defined in the domain of inexact pattern matching~\citep{Navarro2001}, and a common practice is to start with \textit{seeds}, which are exact repeats and then try to extend them to enlarge their support (number of occurrences) while minimizing the added differences. 
Such an idea has been used for instance for DNA compression~\citep{Chen2001,Galle2011}.
The particular kind of inexact motif we focus on is based on insights from theoretical and experimental results in grammatical inference.
We assume that $v$ and $v'$ share a common prefix and suffix, and that all the changes are contained in the middle.
This is, $v = p w s$ and $v' = p w' s$, with $w \neq w'$.
This corresponds to typical distributional approaches which look for words $v,v'$ that occur in the same context, and has been applied as such similarly in ABL~\citep{ABL}

\citet[Section 2.2]{thesisVanzaanen} argues that replacing unequal parts leads to a smaller grammar than replacing equal parts.
However, the analysis there does not take into account that replacing unequal parts adds ambiguity, which -- from a lossless compression perspective -- has to be disambiguated in order to retrieve the correct sequence.
It is surprisingly hard to define an encoding in such a way that replacing such motifs results in grammars that are smaller than those that can be obtained by replacing exact repeats only, as has been reported previously~\citep{Dorr2014}. 
In the remainder of this section we will describe several attempts of finding such encodings.

\subsection{Algorithm}

We propose to extend the greedy algorithm for inferring small straight-line grammars~\citep{Apostolico2000,Nevill2000} to take into account such inexact motifs.
That algorithm (\texttt{Greedy}) chooses in each iteration the repeat that reduces the most the current grammar.
For an exact repeat $u$, the gain $f(u,G)$ is the reduction in size of replacing all occurrences\footnote{Actually, a maximal set of non-overlapping occurrences.} of $u$ in $G$ by a new non-terminal $N$ and adding a rule $N \rightarrow u$.
By encoding the grammar in a single string, separating rules by special symbols\footnote{So that the size of a grammar is just $\displaystyle \sum_{N \rightarrow \alpha \in \Prod}|\alpha|+1$}, it can easily be shown that $f(u,G) = (|u|-1) ~ (\occ_G(u)-1) - 2$, where $\occ_G(u)$ is the number of non-overlapping occurrences of $u$ in $G$.
Deducing such a formula for branching non-terminals is a bit more complicated, and depends strongly on the specific encoding used.

\subsection{Encoding the Grammars}
In order to provide a fair comparison with the straight-line grammars, we will model carefully the way the grammar is encoded.
It should be done in such a way that the target sequence can be retrieved unambiguously from that encoding alone.
This choice will then guide the optimization procedure to minimize it, and as we will see, it influences heavily the resulting grammar.

As a technical point we note that we take advantage of the sequential nature of the final encoding to sort and re-name the non-terminals conveniently. 
Before encoding the final grammar, the non-terminals are sorted by their depth in the parsing trees.
For this, we define $\textit{depth}_G(N)$ for $N \in \N$ as the maximal depth over all parse trees of all sequences $s \in L(G)$, and have the following:

\begin{proposition}
If $G$ is a non-recursive grammar, then $\text{depth}_G(N)$ is well-defined.
\end{proposition}

\noindent which comes directly from the absence of recursion in these grammars.



We first choose to encode all the possible branchings sequentially, separated by a special separator symbol ($|$, where $| \not \in (\Sigma \cup \N)$).

\subsubsection{Variable-length encoding}

In the most general setting, we are interested in motifs of the form $u.^*v$, which match any substring of the form $uwv$ with $w \in (\Sigma + \N)^*$.
Searching for such general motifs is computationally expensive but feasible, by considering all pairs of repeats.

As an example, consider the following sequence:

\begin{tabular}{ll}
s = & 
  Alice was beginning to get very tired  \\ 
& Alice was getting very tired \\
& Alice is very tired  \\
& Alice will be very tired  \\ 
& Alice was getting very tired  \\
\end{tabular}

\smallskip

\noindent where we assume the alphabet to be the set of English words.
Consider now the following grammar $G_1$ generating $s$:

\begin{equation}
\begin{aligned}
S \rightarrow~& N ~ N ~ N ~ N ~ N  \\
O \rightarrow~&\text{Alice~} V \text{~very tired}\\ 
I \rightarrow~&\text{was beginning to get}~|~\text{was getting}~|~\text{is}~|~\text{will be}~|~\text{was getting}
\end{aligned}
\label{exGrammar}
\end{equation}

\noindent which would be encoded as ``$N_1 N_1 N_1 N_1 N_1$ \# Alice $N_2$ is very tired \# was beginning to get $|$ was getting $|$ is $|$ will be $|$ was getting \#'', where $\#$ is an end-of-rule separator and $|$ the choice separator.
Note that the expansion ``was getting'' is repeated twice. 
While this provides redundancy in this case, it is needed to ensure a unique decoding. 
An alternative solution, for instance, would be to provide a list of occurrences and spelling out each unique expansion only once.
However, this does not end up with a better gain in the end, and -- in addition -- if this repetition represents a significant loss (because it occurs many times, or it is very long), it should be captured by a non-branching rule.

For such an encoding, $\textit{cost(s|G)}$ is the same for all $s$ and is included in the encoding of the grammar.
In this example, $|G_1| = 28$.

\smallskip

Unfortunately, this choice of general motifs and encoding proves to be unfit to compete against simple repeats.
It can be shown that the reduction in the grammar size achieved by replacing one such motif is always bounded by the gain obtained by replacing both exact repeats $u$ and $v$.

\subsubsection{Fixed-length encoding}

The reason for the lack of improvement with variable-length encoding is the additional overhead from the separator symbols ($|$).
A standard strategy in data compression to get rid of separators is to focus on fixed-length words.
We adapted this by restricting the inside part of the motif to be of fixed size.
This is, we search for motifs of the form $u.^kv$, which match any substring of the form $uwv$ with $w \in (\Sigma \cup \N)^k$.
While more restrictive in what they can capture, those motifs allow a more efficient encoding.
An example grammar $G_2$ that represents sequence $s$ and uses the knowledge of fixed-length motifs is:
\begin{equation}
\begin{aligned}
S \rightarrow~	&\text{Alice was beginning to get very tired } \\
                & O ~ ~  \text{Alice is very tired }  ~ ~  O ~ O \\
O \rightarrow~&\text{Alice~} I \text{~very tired } \\                
I \rightarrow~&\text{was getting}~|~\text{will be}~|~\text{was getting}
\end{aligned}
\end{equation}
\noindent which would be encoded as `` Alice was beginning to get very tired $N_1$  Alice is very tired $N_1$ $N_1$  \# Alice $N_2$ very tired  \# was getting $|$ will be was getting''.
The separator symbol now has to be used only the first time, indicating the length of the expansion.
As all right-hand sides of the same inside rule now have fixed-length, this information can be used to retrieve all production rules unambigously.
The length of the expansion until that point, together with the number of occurrences of $N_1$ indicates the end of that rule without need of providing an additional end-of-rule symbol.
The total size of $G_2$ is then $27$.
\\\indent We can now deduce the gain introduced by replacing a motif $u.^kv$ with non-terminals $O$ and $I$:

$\begin{array}{rcll}
f(u.^kv,G) = & & \left((|u|+|v|) ~  \occ_G(u.^kv)  \right)  & \text{gain in the sequence} \\
		   & - & \occ_G(u.^kv) & \text{new non-terminal} \\
           & - & \left(|u|+|v|+1+1\right) & \text{$O$ rule + separator symbols} \\
           & - & \left(1 + 1\right) & \text{separator symbols for $I$ rule} \\
          =  & & \\
			 & \multicolumn{3}{l}{\left((|u|+|v|-1) ~ (\occ_G(u.^kv) -1)\right)-5}
\end{array}$


This results in the algorithm \algname given in Alg.~\ref{alg:greedy}.
$G_{w \mapsto N}$ refers to the grammar where all non-overlapping occurrences of the string $w$ are replaced by the new non-terminal $N$ and $N \rightarrow w$ is added to the productions.
Similarly $G_{u.^kv \mapsto O,I}$ is the grammar where all non-overlapping realizations of $u.^kv$ are replaced by the new non-terminal $O$, and the rules are extended with $O \rightarrow u I v$ and $I \rightarrow w$ for all $w$ such that $uwv$ occurs in $G$.

While a smallest possible fixed-length non-recursive grammar is obviously smaller than a smallest straight-line grammar (because more general), our experiments (see Sect.~\ref{sect:experiments}) show that the final grammars obtained with \algname are actually larger than those obtained by minimizing the size of straight-line grammars only.

\subsection{Post-processing Algorithm}
We finally report the results of a simple but effective method that starts from any straight-line grammar, and infers branching rules from it (Alg.~\ref{alg:postalg}).
This is reminiscent of work done to generalize the output of the \texttt{Sequitur} algorithm~\citep[Chapter 5]{NevillThesis}.

The algorithm starts from any proposal for the smallest grammar problem.
We then search for fixed-lengths motifs $u.^kv$, replace them greedily one by one until no further compression can be achieved starting with the one that achieves the highest compression.
Note that, because the algorithm starts from a straight-line grammar with no positive-gain repeats left, all repeated left and right contexts are of length one\footnote{Which could of course be a non-terminal.}, therefore reducing greatly the execution time.
            
\RestyleAlgo{boxruled}
\LinesNumbered
\begin{minipage}[h]{0.5\textwidth}
\vspace{0pt}  
\begin{algorithm}[H]
        \caption{Greedy algorithm \algname to compute small non-recursive grammar generating $s$\label{alg:greedy}}
        \SetAlgoLined
        \KwData{string $s$}
        \KwResult{non-recursive grammar $G$ such that $s \in L(G)$}
        $G \assign \langle \Sigma(s), \{S\}, S, \{S \rightarrow s\} \rangle $\;
        \While{\bf true}{
        	$\displaystyle w \assign \max_{w \in (\Sigma \cup \N)^*} f(w,G) $\;
            $\displaystyle u.^kv \assign \max_{u,v \in (\Sigma \cup \N)^*, k \in \mathbb{N}} f(u.^kv,G) $\;
            \lIf{$f(w,G) \leq 0 \wedge f(u.^kv,G) \leq 0$}{
    \textbf{return} $G$
  }
  \uElseIf{$f(w,G) > f(u.^kv,G)$}{
    $N$ is a fresh non-terminal\;
    $G \assign G_{w \mapsto N}$\;
  }
  \uElse{
  	$O,I$ are fresh non-terminals\;
    $G \assign G_{u.^kv \mapsto O,I}$\;
  }
        }
\end{algorithm}
\end{minipage}
\begin{minipage}[h]{0.5\textwidth}
\vspace{0pt}  
\begin{algorithm}[H]
        \caption{Post-Processing algorithm to compute small non-recursive grammar generating $s$\label{alg:postalg}}
        \SetAlgoLined
        \KwData{string $s$, SGP algorithm \texttt{sgp}}
        \KwResult{non-recursive grammar $G$ such that $s \in L(G)$}
        $G \assign \texttt{sgp}(s)$\;
        \While{\bf true}{
            $\displaystyle u.^kv \assign \max_{u,v \in (\Sigma \cup \N)^*, k \in \mathbb{N}} f(u.^kv,G) $\;
            \uIf{$f(u.^kv,G) \leq 0$}{
			    \textbf{return} $G$\;
			} \uElse{
			  	$O,I$ are fresh non-terminals\;
			    $G \assign G_{u.^kv \mapsto O,I}$\;
		    }
        }
\end{algorithm}
\end{minipage}

\section{Experimental Results}
\label{sect:experiments}
We compared the effectiveness of our proposed algorithm with algorithms that approximate the smallest grammar problem in two areas.
The first is the direct goal of SGP, namely, to find small grammars that encode the data.
We show how the more expressive grammar can lead to consistently smaller grammars, and considerably so in sequences with a large number of fixed-size motifs.
We also report results on qualitative measures of the obtained structure, using a linguistic corpus annotated with its syntactic tree structure.

\subsection{Smaller Grammars}
In this section, we compare the compression performance of our algorithm with four SGP solvers: \texttt{Greedy}~\citep{Nevill2000,Apostolico2000}, IRRMGP, ZZ~\citep{Carrascosa2011b} and \emph{MMAS-GA}~\citep{Benz2013}.
We report the results on two datasets widely used in data compression: a DNA corpus\footnote{\url{http://people.unipmn.it/~manzini/dnacorpus/historical/}} and the Canterbery corpus\footnote{\url{http://corpus.canterbury.ac.nz/}}.
Details about these corpora are in Table~\ref{tb:datasets}.

\begin{table}[t]
\centering
\scalebox{.6}{
\centering
\renewcommand{\arraystretch}{1.5}
\begin{tabular}{l c r r}
\multirow{-2}{*}{\bf sequence}& \multirow{-2}{*}{\bf $|s|$}&\multirow{-2}{*}{$|\Sigma(s)|$}&\multirow{-2}{*}{$|\mathcal{R}(s)|/|s|$}
\\\thickhline
chmpxx & 121,024 &4&0.82
\\
chntxx & 155,844 &4&0.77
\\
hehcmv & 229,354 &4&1.46
\\
humdyst & 38,770 &4&0.77
\\
humghcs & 66,495 &4&13.77
\\
humhbb & 73,308 &4&9.01
\\
humhdab & 58,864 &4&1.21
\\
humprtb & 56,737 &4&1.07
\\
mpomtcg & 186,609 &4&1.36
\\
mtpacga & 100,314 &4&0.97
\\
vaccg & 191,737 &4&2.21
\end{tabular}}
~~~~~~~~~
\scalebox{.6}{
\centering
\renewcommand{\arraystretch}{1.5}
\begin{tabular}{l c r r}
\multirow{-2}{*}{\bf sequence}& \multirow{-2}{*}{\bf $|s|$}&\multirow{-2}{*}{$|\Sigma(s)|$}&\multirow{-2}{*}{$|\mathcal{R}(s)|/|s|$}
\\\thickhline
alice29.txt & 152,089 &74&1.45
\\
asyoulik.txt & 125,179 &68&1.22
\\
cp.html & 24,603 &86&4.32
\\
fields.c & 11,150 &90&5.03
\\
grammar.lsp & 3,721 &76&3.43
\\
kennedy.xls & 1,029,744 &256&0.08
\\
lcet10.txt & 426,754 &84&2.00
\\
plrabn12.txt & 481,861 &81&1.02
\\
ptt5 & 513,216 &159&194.74
\\
sum & 38,240 &255&17.44
\\
xargs.1 & 4,227 &74&1.77

\end{tabular}}

\caption{Statistics of the the DNA corpus (left) and Canterbury (right). We report length, number of different symbols and the number of repeats normalized by length.}
\label{tb:datasets}
\end{table}

\smallskip

\begin{table}[t]
\centering
\scalebox{.65}{
\centering
\renewcommand{\arraystretch}{1.5}
\begin{tabular}{l"r r r r | r r r | r r r | c}
&&\multicolumn{5}{c}{\bf Grammar Size}
\\
\multirow{-2}{*}{\bf Data}&{\bf \texttt{Greedy}}&{\bf \algname} &{\bf +Post}&
$\#Ctx$&{\bf IRRMGP}&{\bf +Post}&
$\#Ctx$&{\bf ZZ}&{\bf +Post}&
$\#Ctx$
&
{\bf MMAS-GA}
\\\thickhline
chmpxx & 28,704 &{\bf 29,477} {\bf \large\color{red}{$\uparrow$}} & {\bf 28,534} {\bf \large\color{ForestGreen}{$\downarrow$}} & 
$64$& 27,683&{\bf 27,584} {\bf \large\color{ForestGreen}{$\downarrow$}} & 
$27$ & 26,024 &{\bf 26,024} {\bf \large\color{gray}{$=$}} & 
$0$&25,882
\\
chntxx & 37,883 &{\bf 38,212} {\bf \large\color{red}{$\uparrow$}} &{{\bf 37,703}} {\bf \large\color{ForestGreen}{$\downarrow$}} & 
$71$&  36,405&{\bf 36,285} {\bf \large\color{ForestGreen}{$\downarrow$}} & 
$25$ & 33,942 &{\bf 33,942} {\bf \large\color{gray}{$=$}} & 
$0$&33,924
\\
hehcmv & 53,694&{\bf 54,451} {\bf \large\color{red}{$\uparrow$}} &{{\bf 53,398}}  {\bf \large\color{ForestGreen}{$\downarrow$}} & 
$113$ &  51,369&{\bf 51,242} {\bf \large\color{ForestGreen}{$\downarrow$}} & 
$30$ & - & - & 
-&48,443
\\
humdyst & 11,064 &{\bf 11,166} {\bf \large\color{red}{$\uparrow$}} & {{\bf 11,017}} {\bf \large\color{ForestGreen}{$\downarrow$}} & 
$19$& 10,700 &{\bf 10,680} {\bf \large\color{ForestGreen}{$\downarrow$}} & 
$6$ & 10,037 &{\bf 10,037} {\bf \large\color{gray}{$=$}} & 
$0$&9,966
\\
humghcs & 12,937 &{\bf 13,655} {\bf \large\color{red}{$\uparrow$}} &{\bf 12,908} {\bf \large\color{ForestGreen}{$\downarrow$}} & 
$14$&12,708  &{\bf 12,708} {\bf \large\color{gray}{$=$}} & 
$0$ & 12,033 &{\bf 12,023} {\bf \large\color{ForestGreen}{$\downarrow$}} & 
$5$&12,013
\\
humhbb & 18,703 &{\bf 18,893} {\bf \large\color{red}{$\uparrow$}} &{{\bf 18,614}} {\bf \large\color{ForestGreen}{$\downarrow$}} & 
$36$& 18,133 &{\bf 18,060} {\bf \large\color{ForestGreen}{$\downarrow$}} & 
$20$ & 17,026 &{\bf 17,024} {\bf \large\color{ForestGreen}{$\downarrow$}} & 
$1$&17,007
\\
humhdab & 15,311& {\bf 19,736} {\bf \large\color{red}{$\uparrow$}} & {{\bf 15,242}} {\bf \large\color{ForestGreen}{$\downarrow$}} & 
$27$ &  14,906&{\bf 14,879} {\bf \large\color{ForestGreen}{$\downarrow$}} & 
$8$ & 13,995 &{\bf 13,995} {\bf \large\color{gray}{$=$}} & 
$0$&13,864
\\
humprtb & 14,884&{\bf 17,122} {\bf \large\color{red}{$\uparrow$}} &{{\bf 14,817}} {\bf \large\color{ForestGreen}{$\downarrow$}} & 
$26$ &  14,492&{\bf 14,451} {\bf \large\color{ForestGreen}{$\downarrow$}} & 
$11$ & 13,661 &{\bf 13,661} {\bf \large\color{gray}{$=$}} &
$0$&13,528
\\
mpomtcg & 44,175&{\bf 45,018} {\bf \large\color{red}{$\uparrow$}} &{{\bf 43,930}} {\bf \large\color{ForestGreen}{$\downarrow$}} & 
$89$ &  42,825&{\bf 42,658} {\bf \large\color{ForestGreen}{$\downarrow$}} & 
$40$ & 39,913 &{\bf 39,911} {\bf \large\color{ForestGreen}{$\downarrow$}} & 
$1$&39,988
\\
mtpacga & 24,556 & {\bf 24,878} {\bf \large\color{red}{$\uparrow$}} &{{\bf 24,408}} {\bf \large\color{ForestGreen}{$\downarrow$}} & 
$58$&  23,682&{\bf 23,608} {\bf \large\color{ForestGreen}{$\downarrow$}} & 
$16$ & 22,189 &{\bf 22,189} {\bf \large\color{gray}{$=$}} & 
$0$&22,072
\\
vaccg & 43,711&{\bf 44,261} {\bf \large\color{red}{$\uparrow$}} &{{\bf 43,445}} {\bf \large\color{ForestGreen}{$\downarrow$}} & 
$99$ &  41,882&{\bf 41,778} {\bf \large\color{ForestGreen}{$\downarrow$}} & 
$29$ & - &- & 
-&39,369
\\\thickhline
alice29.txt & 41,001 &{\bf 50,777} {\bf \large\color{red}{$\uparrow$}} &{{\bf 40,984}} {\bf \large\color{ForestGreen}{$\downarrow$}} & 
$7$& 40,218 &{\bf 40,218} {\bf \large\color{gray}{$=$}} & 
0 & 37,702 &{\bf 37,662} {\bf \large\color{ForestGreen}{$\downarrow$}} & 
$12$&37,688
\\
asyoulik.txt & 37,475&{\bf 45,520} {\bf \large\color{red}{$\uparrow$}} &{{\bf 37,464}} {\bf \large\color{ForestGreen}{$\downarrow$}} & 
$4$ & 36,910 &{\bf 36,905} {\bf \large\color{ForestGreen}{$\downarrow$}} & 
$1$ & 35,001 &{\bf 34,953} {\bf \large\color{ForestGreen}{$\downarrow$}} & 
$16$&34,967
\\
cp.html & 8,049 &{\bf 8,310} {\bf \large\color{red}{$\uparrow$}} &{{\bf 8,003}} {\bf \large\color{ForestGreen}{$\downarrow$}} & 
$6$&7,974 &{\bf 7,971} {\bf \large\color{ForestGreen}{$\downarrow$}} & 
$1$ & 7,768 &{\bf 7,747} {\bf \large\color{ForestGreen}{$\downarrow$}} & 
$9$&7,746
\\
fields.c & 3,417 &{\bf 3,681} {\bf \large\color{red}{$\uparrow$}} &{{\bf 3,380}} {\bf \large\color{ForestGreen}{$\downarrow$}} & 
$7$&  3,385&{\bf 3,381} {\bf \large\color{ForestGreen}{$\downarrow$}} & 
$1$ & 3,312 &{\bf 3,285} {\bf \large\color{ForestGreen}{$\downarrow$}} & 
$13$&3,301
\\
grammar.lsp & 1,474&{\bf 1,475} {\bf \large\color{red}{$\uparrow$}} &{{\bf 1,458}} {\bf \large\color{ForestGreen}{$\downarrow$}} & 
$2$ &  1,472&{\bf 1,472} {\bf \large\color{gray}{$=$}} & 
$0$ & 1,466 &{\bf 1,462} {\bf \large\color{ForestGreen}{$\downarrow$}} & 
$1$&1,452
\\
kennedy.xls & 166,925 & - & {\bf 99,915} {\bf \large\color{ForestGreen}{$\downarrow$}} & 
1,233 & 166,810 & {\bf 98,479} {\bf \large\color{ForestGreen}{$\downarrow$}} & 
1,174 & 166,705 & {\bf 98,258} {\bf \large\color{ForestGreen}{$\downarrow$}} & 
1,161&166,534
\\
lcet10.txt & 90,100 &{\bf 115,625} {\bf \large\color{red}{$\uparrow$}}&{{\bf 89,998}} {\bf \large\color{ForestGreen}{$\downarrow$}} & 
$33$& 88,778 &{\bf 88,750} {\bf \large\color{ForestGreen}{$\downarrow$}} & 
$9$ & - & - & 
-&87,086
\\
plrabn12.txt & 124,199&{\bf 165,122} {\bf \large\color{red}{$\uparrow$}} &{{\bf 124,009}} {\bf \large\color{ForestGreen}{$\downarrow$}} & 
$58$ &120,770 &{\bf 120,760} {\bf \large\color{ForestGreen}{$\downarrow$}} & 
$2$ & - & - & 
-&114,960
\\
ptt5 & 45,135 & - &{\bf 45,118} {\bf \large\color{ForestGreen}{$\downarrow$}} & 
$7$ & 44,129 & {\bf 44,123} {\bf \large\color{ForestGreen}{$\downarrow$}} & 
3 &-&-&
-&42,661
\\
sum & 12,207 &{\bf 14,722} {\bf \large\color{red}{$\uparrow$}} &{{\bf 11,761}} {\bf \large\color{ForestGreen}{$\downarrow$}} & 
$52$&12,127  &{\bf 11,868} {\bf \large\color{ForestGreen}{$\downarrow$}} & 
$34$ & - & - & 
-&12,009
\\
xargs.1 & 2,006&{\bf 2,092} {\bf \large\color{red}{$\uparrow$}} &{\bf 2,006} {\bf \large\color{gray}{$=$}} & 
$0$ &  1,993&{\bf 1,990} {\bf \large\color{ForestGreen}{$\downarrow$}} & 
$1$ & 1,973 & {\bf 1,948} {\bf \large\color{ForestGreen}{$\downarrow$}} & 
$3$&1,955
\end{tabular}
}
\caption{Size of the final grammars obtained with the different algorithms. SGP algorithms (non-bold numbers) generate straight-line grammars, while \algname and the \textbf{+Post} columns (bold numbers) infer non-recursive grammars.
Green/Red down-/up-ward arrows show a reduction/increase in the grammar size with respect to the output of the reference algorithm in each section, and \emph{=} shows no change. 
$\#Ctx$ is the number of branching rules detected by the post-processing algorithm. In the cases with -, either the final grammar or the output of ZZ algorithm was not available. The results for MMAS-GA are taken from \citet{Benz2013}.}
\label{tb:size}
\end{table}

The results on the final sizes are reported in Table~\ref{tb:size}.
As pointed out before, we did not achieve to obtain smaller grammars by incorporating branching-rules inference during the main process (algorithm \algname).
The final grammars were consistently larger than the simplest baseline (\texttt{Greedy}), often considerably so (see for instance \texttt{humhdab}, \texttt{alice29.txt}).
However, the same idea of inferring fixed-motifs proved to be successful when applied as a post-processing.
Moreover, this strategy can be applied to any straight-line grammar and can therefore be used after any SGP algorithm.
Under the column \#Ctx, we give the number of branching rules that are inferred.
The number of occurrences of these rules is of course much higher in general.

While the reduction in size is small, it applies consistently throughout all the SGP algorithms we tried\footnote{We did not have access to the final grammars of \emph{MMAS-GA}}.
The better the original algorithm, the smaller the gain.
While this may point towards a convergence of the possible redundancy that can be extracted, it should be noted that our approach runs much faster than the more sophisticated algorithms (\emph{ZZ}, \emph{MMAS-GA}).
Moreover, our best result in Table~\ref{tb:size} become the new state-of-the-art in several cases, and we would expect an even better improvement if starting from the final grammars output by \emph{MMAS-GA}.

We analyzed separately the huge difference in the gain obtained on the \texttt{kennedy.xlsx} file.
This is a binary file, encoding a large spreadsheet ($347 \times 228$, in Excel format) containing numerical values, many of which are empty.
Most of the gains over the straight-line baselines seem to come from the way these numbers are getting encoded, with a common prefix and suffix and a fixed-length field for the specific value.
These fields are therefore ideal candidates for our branching-rule inference. 
We were able to recreate those results by generating random Excel tables, obtaining improvements of 6 to 33\% (relative to the original size of IRRMGP) depending on the number of non-zero entries the table had.

\subsection{Better Structure}
Following the original motivation for closing the gap between the structures found in SGP and the structures that are sought in grammatical inference, we evaluated the obtained branching rules by their capacity for unsupervised parsing.
For this we benchmarked our method in the task of \textit{unsupervised parsing}, the problem of retrieving the correct tree structure of a natural language text.
We took the standard approach in the field, starting from the Part-of-Speech (POS) tags of the Penn Tree bank dataset \citep{Marcus1994}.
Current \textit{supervised} methods achieve a performance above $0.9$ of $F_1$ measure~\citep{Vinyals2015}.
As expected, \textit{unsupervised} approaches report worse performance, around $0.8$~\citep{Scicluna2014}. 
These are in general very computationally intense methods and performance is only reported on top of WSJ10, sentences of size up to $10$. 
We diverge from that, reporting results on all \numprint{49208} sentences\footnote{Excluding sentence $1855$, for which the EVALB evaluation tool we used had trouble processing.}.
For evaluation, we used precision over the set of brackets, together with the percentage of non-crossing brackets~\citep{Klein2005}, a standard practice for which we relied on the EVALB tool\footnote{\url{nlp.cs.nyu.edu/evalb/}} which removes singleton and sentence-wide brackets.
While precision is the percentage of correctly retrieved brackets, non-crossing brackets is the percentage over the retrieved brackets that do not contradict a gold bracket and give an idea on how not-incorrect the results are (as opposed to correct).

Our focus is on comparing the quality of the brackets of the branching rules with those of the non-branching rules.
We furthermore distinguish the brackets covering a context, and the one covering an inside.
For the rules $\{O \rightarrow \alpha I \beta, I \rightarrow \gamma_1 | \gamma_2\}$, the inside brackets cover $\gamma_1$ and $\gamma_2$, while the context rules cover $\alpha \gamma_1 \beta$ and $\alpha \gamma_2 \beta$.
The number of context brackets is always the same as the number of inside brackets.

\smallskip

\begin{figure}[t]
    \centering
    \begin{subfigure}[t]{0.5\textwidth}
        \centering
        \includegraphics[width=\textwidth]{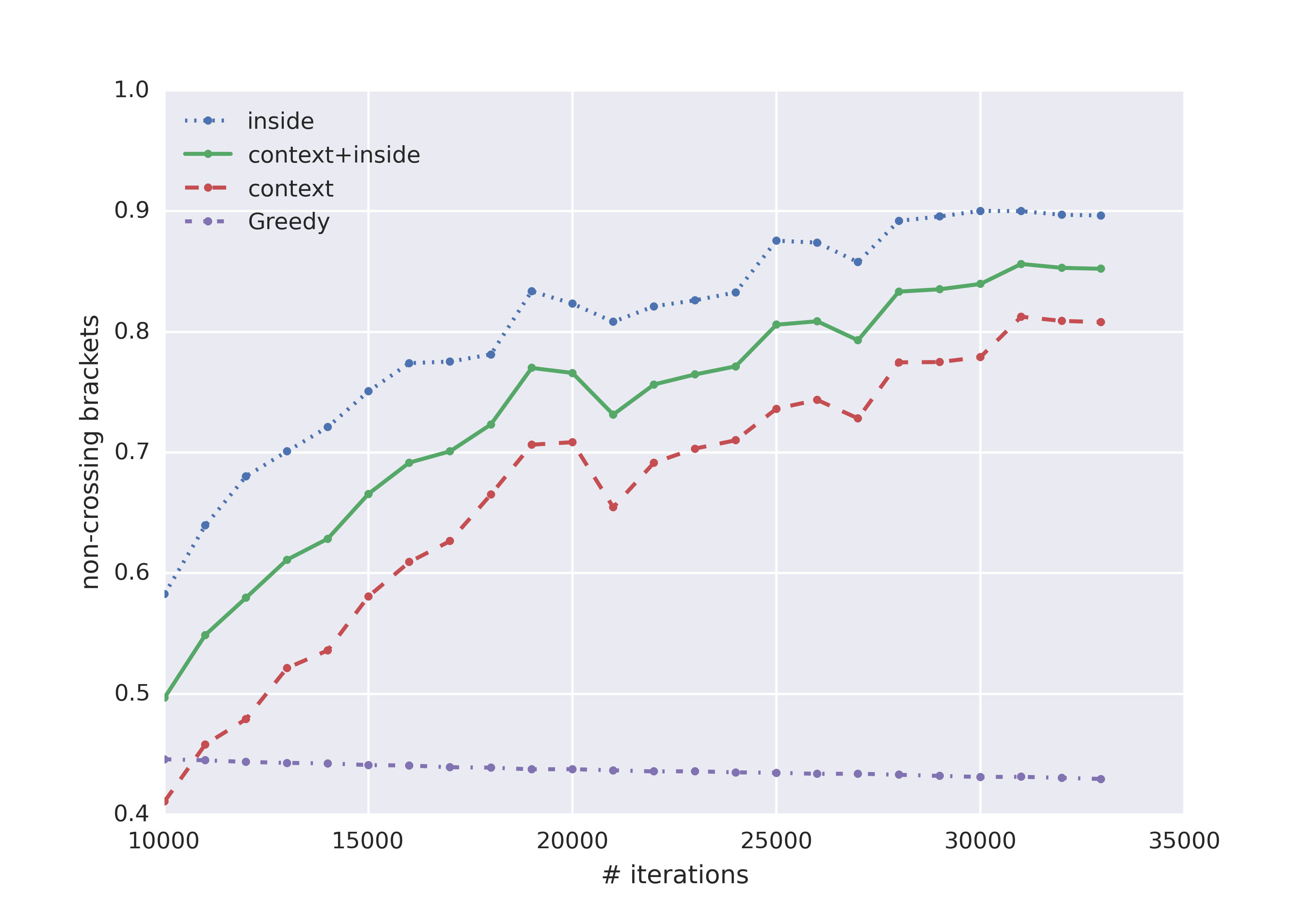}
        \caption{Percentage of non-crossing brackets}
                \label{fig:ncross}
    \end{subfigure}%
    ~ 
    \begin{subfigure}[t]{0.5\textwidth}
        \centering
        \includegraphics[width=\textwidth]{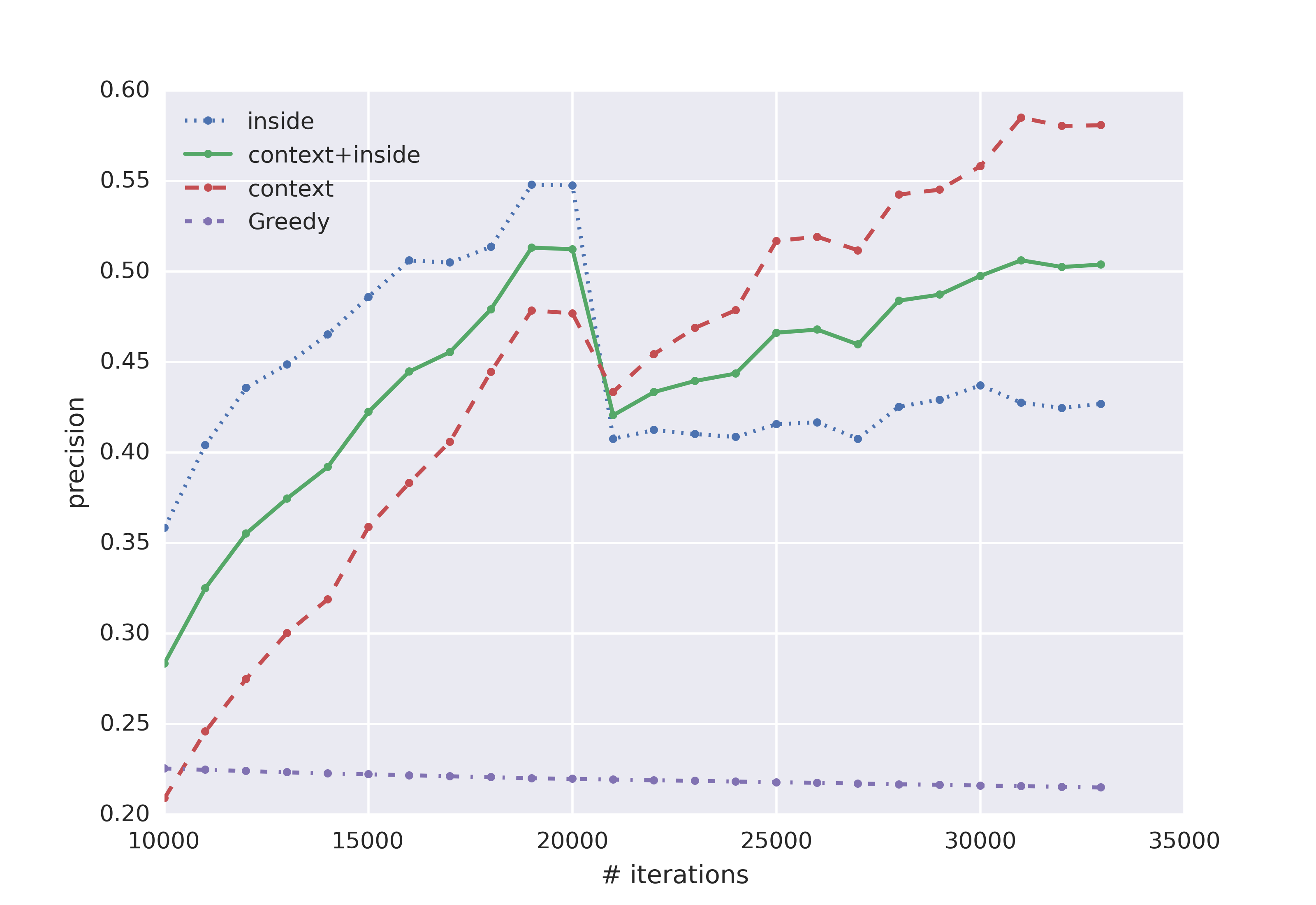}
        \caption{Bracketing precision}
        \label{fig:prec}
    \end{subfigure}
    
	\begin{subfigure}[t]{0.5\textwidth}
        \centering
        \includegraphics[width=\textwidth]{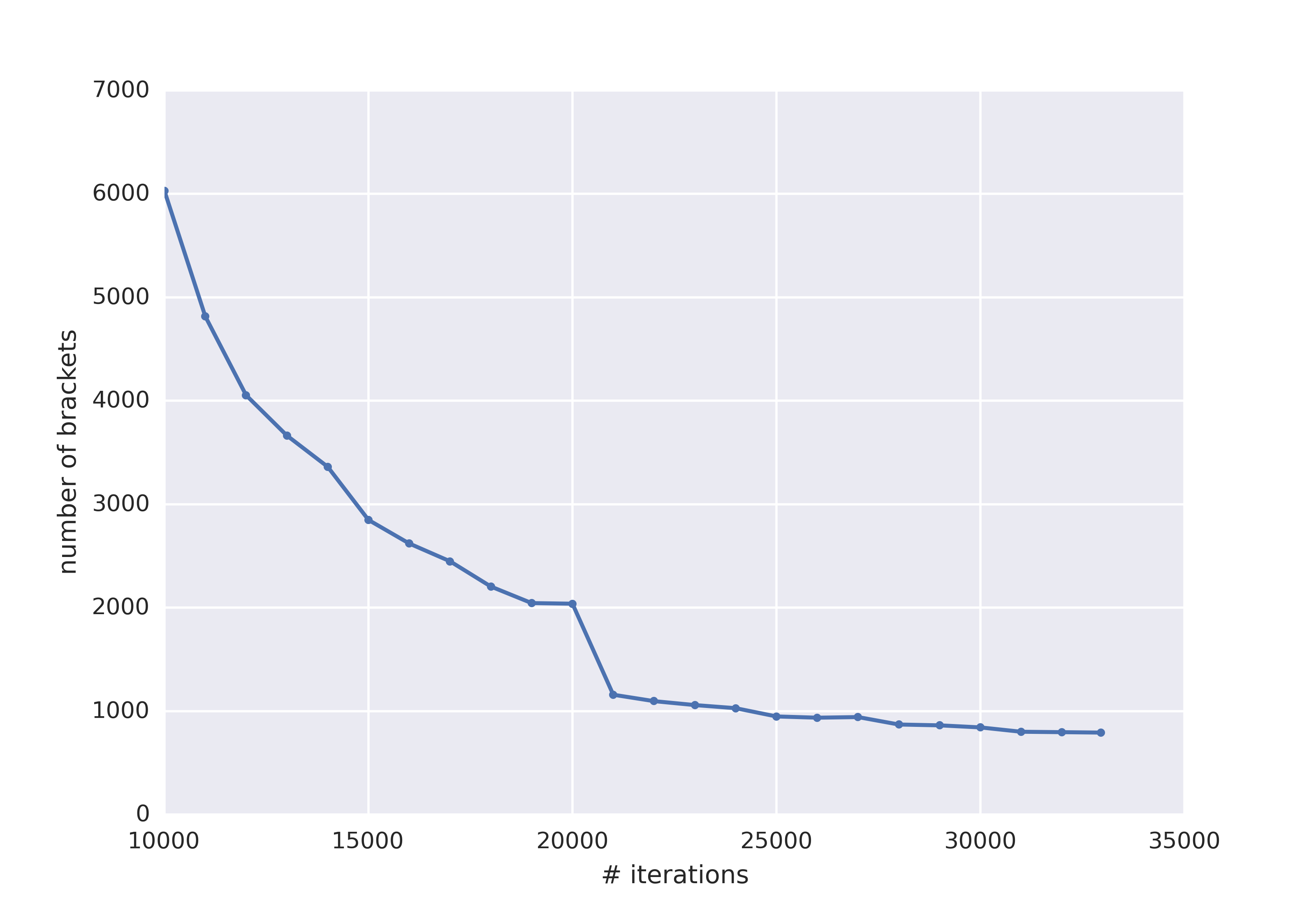}
        \caption{Number of brackets of branching rules.}
        \label{fig:number}
    \end{subfigure}
    
    \caption{Structuring accuracy of the proposed post-processing algorithm. $x$-axis is the number of iterations when we stopped the \texttt{Greedy} algorithm.}        \label{structExperiment}
	\label{fig:parsing}
\end{figure}

As before, we are mainly interested in comparing the additional rules added by non-recursive grammars.
The \texttt{Greedy} algorithm creates around \numprint{950000} brackets, of which only $21\%$ are correct.
The proposed post-processing adds another 792 brackets, but with a much higher precision ($50.48\%$) and mostly consistent ($85\%$).
In order to evaluate the sensitivity of these results, and to see if they generalize if more brackets are retrieved, we stopped the \texttt{Greedy} algorithm earlier: this creates larger grammars, with more options for the creation of branching rules in the post-processing stage.
The final results are summarized in Fig.~\ref{fig:parsing}.
Because of the small number of brackets (Fig.~\ref{fig:number}) we do not report recall.
While the numbers of correct and consistent brackets decreases with increasing number of branching rules, they do so very gently and are much higher than the accuracy of non-branching rules.
Furthermore, a stark difference appears between context and inside brackets: while context brackets are much more often correct (reaching almost 60\%), they are less consistent than inside brackets (which have a non-crossing percentage of 90\%).
These results get their whole meaning when compared to the brackets obtained by just considering the straight-line grammars.
Their accuracy varies very little over the iterations, and is always extremely low (around $22\%$).

Finally, the drop around \numprint{20000} iterations belongs to a point where highly frequent context patterns\footnote{Most notably opening/closing quotation and parentheses} stop being captured by branching rules and are modeled by repeats.
This also means that the good performance at the end is not due to these easy to model constituents.


\smallskip

As said, reported values on unsupervised parsing of this dataset focuses on sentences of length up to 10, which only represents less than $10\%$ of the total corpus.
On these sentences, the context brackets obtain a precision of $80\%$, considerably higher than other reported results, although this is not a fair comparison as the number of retrieved brackets is low.
But it is worth to highlight that parsing the longer sentences did not pose any problems at all in our (not optimized) implementations: in fact the algorithm was run on the concatenation of the overall set (over $1.3M$ tokens).

\section{Conclusions}
In this paper we provide a first step towards applying the results around the Smallest Grammar Problem for grammatical inference.
We identify a probable reason for past failures, and show how to extend the work inferring small straight-line grammars towards non-recursive ones.
Our starting point is the MDL principle, and faithful to this principle, we pay careful attention to the encoding of the final grammars to the point that the final search space is strongly constrained by the chosen encoding.
This allows us to make a fair comparison with SGP algorithms on standard benchmarks used for that problem, as in both cases we allow to retrieve the original sequence unambiguously.
Those algorithms consistently improve over the current state-of-the-art, substantially so in one case (\texttt{kennedy.xls} sequence).
One direction of future work could focus on formalizing the phenomena exhibited by that sequence and where else it occurs.

With respect to the original motivation of structuring the sequences, the additional rules of our algorithm have a much higher precision than other methods for unsupervised parsing, without explicitly trying to optimize for it.
Recall, however, is much lower as very few rules are actually inferred. 
Nevertheless, we believe that our results show the potential of this generalized smallest grammar problem for this task and we are considering on how to build on the efficient algorithms developed in the field to capture more such rules.






\bibliography{icgi}

\end{document}